# DSU-Net：An Improved U-Net Model Based on DINOv2 and SAM2 with Multi-scale Cross-model Feature Enhancement

Yimin Xu [1,2]
1.*Chengdu Institute of Computer Application Chinese Academy of Sciences*
Chengdu, China
2.*University of Chinese Academy of Sciences*
Beijing, China
xuyimin23@mails.ucas.ac.cn

Fan Yang[1,2] *
1.*Chengdu Institute of Computer Application Chinese Academy of Sciences*
Chengdu, China
2.*University of Chinese Academy of Sciences*
Beijing, China
yangfan@casit.com.cn
*Corresponding author

Bin Xu[1,2]
1.*Chengdu Institute of Computer Application Chinese Academy of Sciences*
Chengdu, China
2.*University of Chinese Academy of Sciences*
Beijing, China
xubin23@mails.ucas.ac.cn

***Abstract*—This paper addresses two critical challenges in applying general image segmentation models to specific domains: high training costs and inadequate feature representation. We propose a novel multi-scale feature collaboration framework that leverages DINOv2 and SAM2. Our key innovations are threefold: First, we propose a multi-level feature collaboration mechanism that integrates the DINOv2 with SAM2, enabling effective multi-scale feature fusion enhanced by high-dimensional semantics. Next, we design a lightweight adapter that operates under parameter-freezing conditions, complemented by a cross-layer feature interaction module that facilitates efficient domain knowledge transfer. Finally, we develop a U-shaped network based on U-Net, that performs multi-granularity feature aggregation decoding through spatial feature fusion, and mitigates multi-level accuracy loss using a weighted summation of diverse layer loss functions. Experimental results demonstrate that our method significantly outperforms existing state-of-the-art techniques in Salient Object Detection (SOD) and Camouflaged Object Detection (COD), achieving improvements of 1.25% and 1.02%, respectively. Notably, our approach requires training only 1.38% of the parameters to adapt to various downstream image segmentation tasks, eliminating the need for fine-tuning base model. This framework effectively overcomes the adaptation challenges faced by general segmentation models in specialized contexts, providing a robust solution for image segmentation in critical fields such as medical image analysis and industrial inspection. Project page: https://github.com/CheneyXuYiMin/SAM2DINO-Seg.**

## I. INTRODUCTION

Image segmentation, as a core task in computer vision, aims to divide an image into regions with semantic meaning. Its core includes semantic segmentation (assigning category labels to each pixel) and instance segmentation (further distinguishing different instances within the same category). In recent years, breakthroughs in deep learning technology have made image segmentation crucial in diverse application domains, such as medical image analysis (e.g., tumor boundary localization, organ segmentation) and autonomous driving (e.g., road scene understanding, obstacle detection), becoming an critical technical component for the practical application of artificial intelligence.

Meta's SAM[1] series and DINOv2[2] visual foundation models, with their robust model architecture and extensive pre-training data, have demonstrated exceptional performance in the field of general image segmentation, offering significant potential for the advancement of downstream tasks in image segmentation. However, current methods continue to face several challenges. First, as foundational models, SAM2 and DINOv2 exhibit suboptimal performance in various subdomains such as salient object segmentation and camouflaged object segmentation without manual prompts. The primary causes are feature sparsity and the specificity of category samples, which often result in errors in foreground background segmentation and lead to segmented regions that are unrelated to the specified categories in the absence of human intervention. This phenomenon suggests that relying solely on foundational models is inadequate for downstream tasks like image segmentation across diverse domains. Investigating how to better adapt these models to specific categories and tasks in image segmentation, in order to enhance their professionalism and adaptability in application scenarios, presents significant research value.

To overcome the challenges of image segmentation in specialized downstream tasks using large-scale pre-trained foundation models, this paper proposes DSU-Net, an enhanced UNet model that leverages DINOv2-generated features to guide SAM2 in multi-scale feature collaboration. The main contributions of this paper are as follows:

- To reduce training costs, an adapter has been designed. By introducing a small number of parameters, the model can alleviate the domain differences between the training dataset and the pre training model dataset, and lower the hardware resource threshold required for training.
- A new architecture for dual encoder collaboration has been proposed, which utilizes the advantages of DINOv2 self supervised training to compensate for the biases and illusions that occur in the feature extraction stage of the SAM model. The feature representations of the two encoders are fused to obtain richer semantic information.

- Combining spatial, channel, and pixel attention to achieve adaptive aggregation and decoding of multi granularity features. The effectiveness and universality of our method in detecting salient and disguised objects in image segmentation have been verified through testing on multiple public datasets.

## II. Related work

### *A. U-Net Network and Its Improvements*

U-Net[3] is an encoder-decoder architecture specifically designed for medical image segmentation. It addresses the segmentation challenges in small sample datasets by integrating multi-scale features through skip connections. Its upsampling path transmits contextual information across multiple feature channels. When combined with a data augmentation strategy involving random elastic deformations, it significantly enhances the robustness of medical image segmentation. The original U-Net performs well in tasks such as cell segmentation, but has limited adaptability to complex scenarios, including multi-scale targets and noise interference. To address this, researchers have proposed various improved models. Zhou et al.[4] integrated multi-scale feature fusion into the U-Net, capturing features at different scales through multiple branches. Jha et al.[5] integrated residual modules, channel compression excitation mechanisms, and spatial pyramid pooling to achieve more efficient and precise semantic segmentation. With the rapid advancement of Transformers in natural language processing, Chen et al.[6] integrated the Transformer architecture, which emphasizes global information, with CNNs that capture low-level image features for hybrid encoding, significantly enhancing U-Net's segmentation performance. Cao et al.[7], aiming to address the continued reliance on CNNs in TransUNet, proposed using the Swin Transformer[8] as the backbone for the U-Net encoder, creating the first purely Transformer-based semantic segmentation model.

### *B. Segmentation Anything Model*

The Segmentation Anything Model (SAM) is a groundbreaking image segmentation model developed by Meta AI, distinguished by its promptable universal segmentation capability. It can rapidly identify and segment any object in an image through simple inputs such as clicks, bounding boxes, and text. Since the release of SAM in 2023 and the release of SAM2 in 2024, it rapidly became a research hotspot in both academia and industry, marking a significant milestone in the field of computer vision. Research leveraging SAM as a base model has yielded promising results in various specialized domains. Chen et al.[9] proposed adding an adapter to the SAM model, enabling it to adapt to downstream segmentation tasks across various domains. Yang et al. integrated the SAM model with the U-Net architecture and trained it on a large medical segmentation dataset, achieving zero-shot medical image segmentation and creating a pre-trained model for general medical image segmentation. Xiong et al.[10] proposed using Hiera from SAM2 as the U-Net encoder, achieving state-of-the-art performance in multiple downstream tasks.

### *C. DINO Model*

DINO[2, 11] (Distillation and NO labels) is a self-supervised visual learning model proposed by Facebook AI in 2021. At its core, it is based on the Transformer architecture and achieves unsupervised feature learning through knowledge distillation. As a result, compared to supervised learning, the features in DINO's latent space are highly informative, enabling it to discern fine-grained differences between objects, which is crucial for image segmentation tasks.

## III. Method

The traditional U-Net model uses architectures like ResNet and ViT as visual feature extractors, which either lack the ability to capture multi-scale features or have a large number of parameters, making training challenging. The Hiera module in the SAM2 model with its hierarchical structure and pre-training on extensive segmentation datasets, is highly suitable as a feature extractor for the U-Net segmentation network. Experiments using SAM2 Hiera as the backbone to extract image features (shown in Fig. 1), reveal that the features extracted by Hiera perform well in low-dimensional PCA visualization but poorly in high dimensions. T-SNE visualization further indicates that feature-level clustering performs better in low dimensions and worse in high dimensions. In contrast, the pre-trained Vision Transformer (ViT) network in the DINOv2 model demonstrates superior aggregation of high-dimensional features during extraction. This is evident in PCA dimensionality reduction visualizations, which reveal significantly improved performance in the features it extracts. Furthermore, as DINOv2 is trained using self-supervised learning, it generates generic visual features that are applicable to diverse image distributions and various tasks. These features can be effectively utilized across a wide range of downstream tasks without requiring further fine-tuning (shown in Fig. 2). Building on this observation, we propose a DINOv2 high-dimensional feature-enhanced SAM2 feature collaborative fusion model. The model architecture is depicted in Fig. 3.

DSU-Net employs the SAM2 Hiera module and the DINOv2 ViT module as feature extraction components, incorporating multiple feature fusion modules and channel transformation mechanisms. The overall design follows the U-Net symmetric encoder-decoder architecture.

**Encoder**. In DSU-Net, the Hiera module from SAM2 serves as the backbone of the feature extractor. The feature map output from the 24th layer block of the ViT module in DINOv2 is injected as external knowledge into the final feature map extracted by Hiera, guiding the synergy of Hiera features from high-level semantic features to compensate for the loss of semantics and detail in its higher-level features. It is worth noting that this network freezes the pre-trained parameters of SAM2 Hiera and DINOv2 ViT, efficiently utilizing open-source models to reduce training costs. The input image is scaled to $I \in \mathbb{R}^{3\times352\times352}$ before entering the Hiera module. After passing through the Hiera module, it outputs four feature maps at different levels, ranging from low to high , as $Si \in \{\mathbb{R}^{144\times88\times88}, \mathbb{R}^{288\times44\times44}, \mathbb{R}^{576\times22\times22}, \mathbb{R}^{1152\times11\times11}\}$. Before entering the DINOv2 ViT module, the image is scaled to $I \in \mathbb{R}^{3\times518\times518}$, and after passing through the final Block, the output feature map is $V \in$ $\mathbb{R}^{1024\times37\times37}$. The Hiera and ViT modules used above are the pre-trained Hiera-L and ViTl14, respectively.

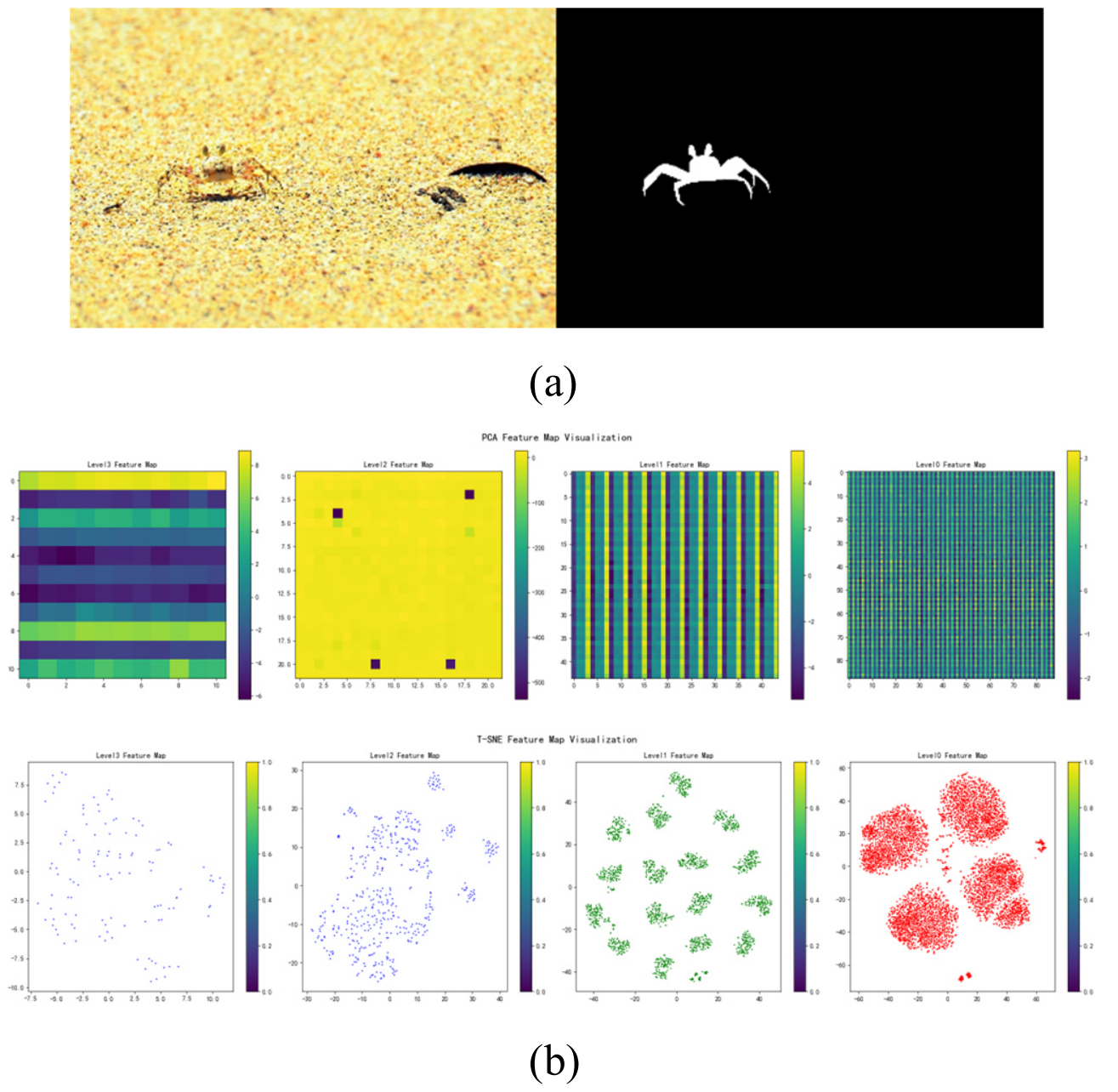

Fig. 1. (a) Input image and its corresponding groundtruth (b) Visualization of features extracted by SAM2 Hiera at different levels using PCA and T-SNE dimensionality reduction

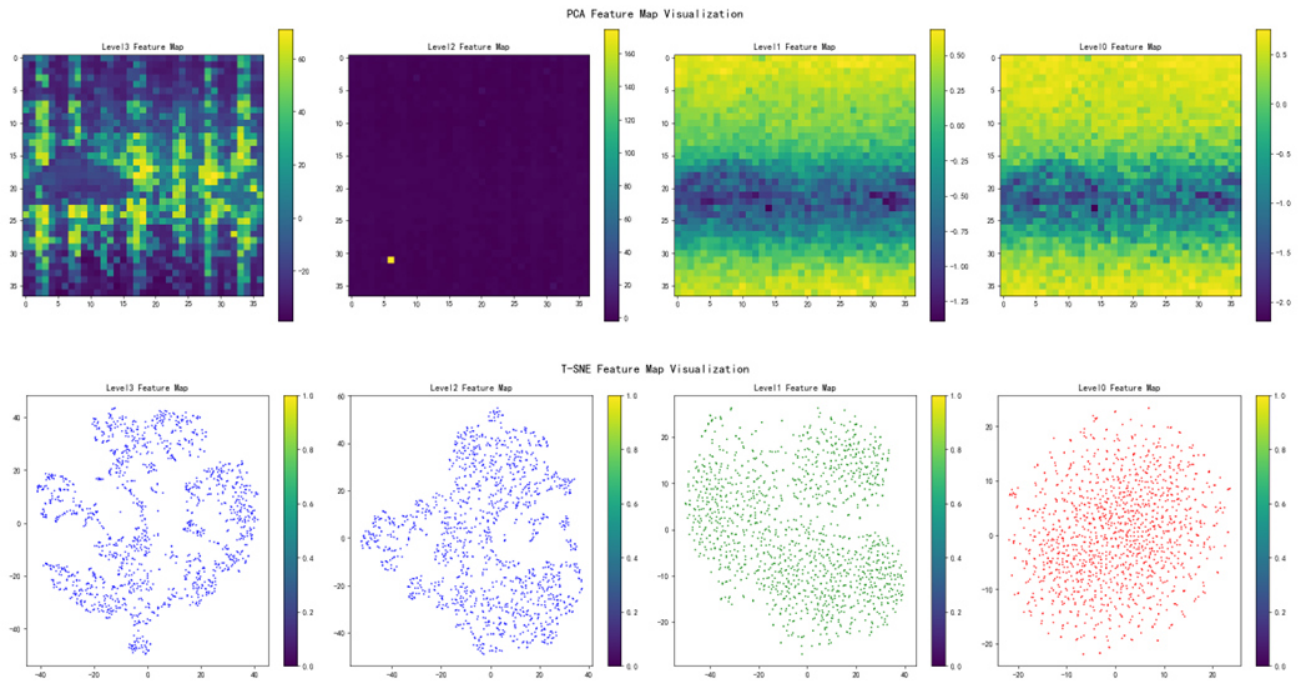

Fig. 2. DINOv2 ViT Feature Extraction at Different Levels with PCA and T-SNE Dimensionality Reduction Visualization

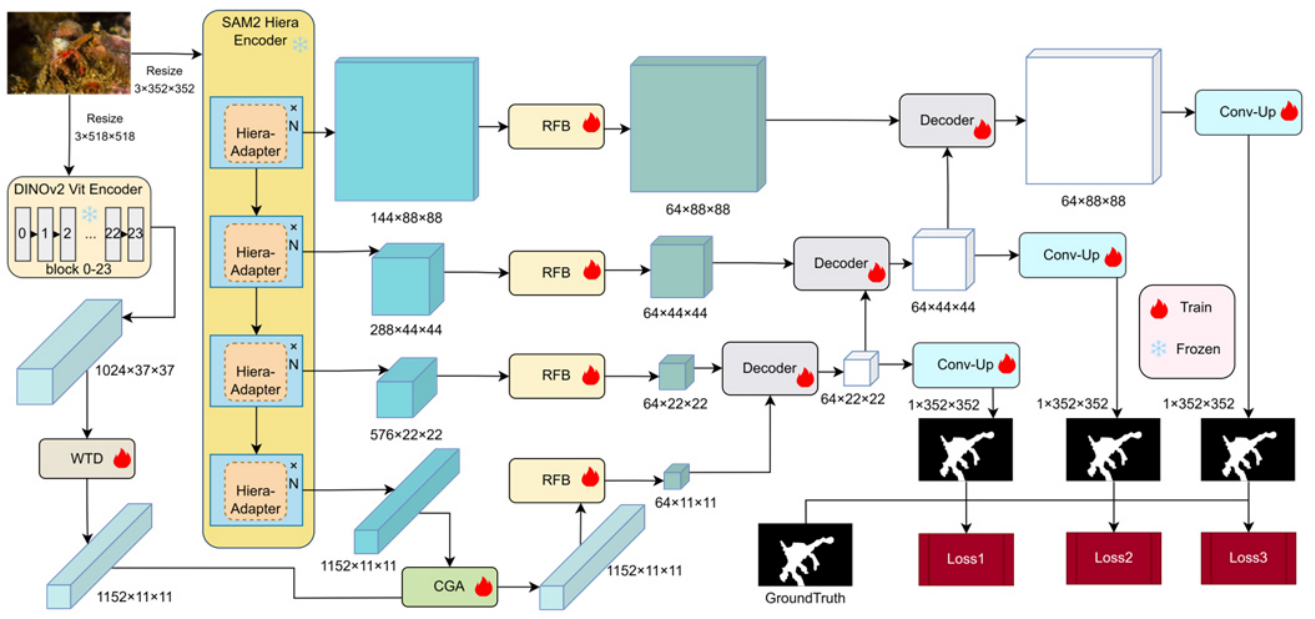

Fig. 3. DSU-Net Model Architecture Diagram.

**Adapter**. The Hiera module contains an extremely large number of parameters, resulting in prohibitively high training time and computational costs. To leverage the powerful feature extraction capabilities of pre-trained models, we introduce the Adapter module, inspired by SAM-Adapter[9] and SAM2-Unet[10]. It first undergoes downsampling through a linear layer, followed by the GeLU activation function, then upsampling through another linear layer, with the GeLU activation function restoring the original dimensionality. The purpose of the adapter is to mitigate the domain differences between the training dataset and the pre-trained model dataset by introducing a small number of parameters, adapting to new dataset scenarios.

**Channel Modulation**. Since the feature map extracted by ViT is $V \in \mathbb{R}^{1024\times37\times37}$, to match the dimensions of the feature map S4 extracted by Hiera ( $S4 \in \mathbb{R}^{1152\times11\times11}$), bilinear interpolation downsampling is first applied to reduce the channel dimension to 1152. Next, depth-wise separable wavelet convolution[12] is employed to reduce both the height and width of the feature map to 11. This is implemented by the WTD module shown in Figure 3, which achieves a larger receptive field with fewer parameters through wavelet transformation. The RFB[13, 14] module is then used to reduce the channel dimension, decreasing the number of channels in the multi-level feature maps extracted by the encoder to 64, thereby enhancing the discriminability and robustness of the feature channels.

**Feature Fusion**. We employ Content-Guided Attention[15] (CGA) to inject the semantic information from features extracted by ViT into those extracted by SAM2 Hiera. By integrating spatial, channel, and pixel attention mechanisms, we fuse the feature representations of both encoders to obtain richer semantic information.

**Decoder**. To utilize high-level feature maps in guiding the fusion of low-level feature maps, preserving spatial detail while enhancing semantic information, we employ the Spatial Feature Fusion module[16] (SFF). This module dynamically adjust the weights of the output feature maps based on the correlations between feature maps at different scales. Finally, 1×1 point-wise convolution and bilinear interpolation upsampling transform the feature maps into $D_i \in \mathbb{R}^{1\times352\times352}$, producing a single-channel image.

**Loss Function**. We use a weighted Intersection over Union (IoU) loss and Binary Cross-Entropy (BCE) loss to quantify the discrepancy between predicted and actual values[10, 13, 17], formulated as:

$$L = L^{W}{}_{IoU} + L^{W}{}_{BCE} \tag{1}$$

Additionally, we compute these loss functions for the decoder output $D_i$ at each level. Thus, the overall loss function for DSU-Net formulated as:

$$L_{total} = w_1 L_1 + w_2 L_2 + w_3 L_3 \tag{2}$$

Among it, $w_1 = 0.25$, $w_2 = 0.5$, $w_3 = 1$.

For a clearer understanding of the structure of each module, detailed diagrams of the aforementioned modules can be found in Fig. 4.

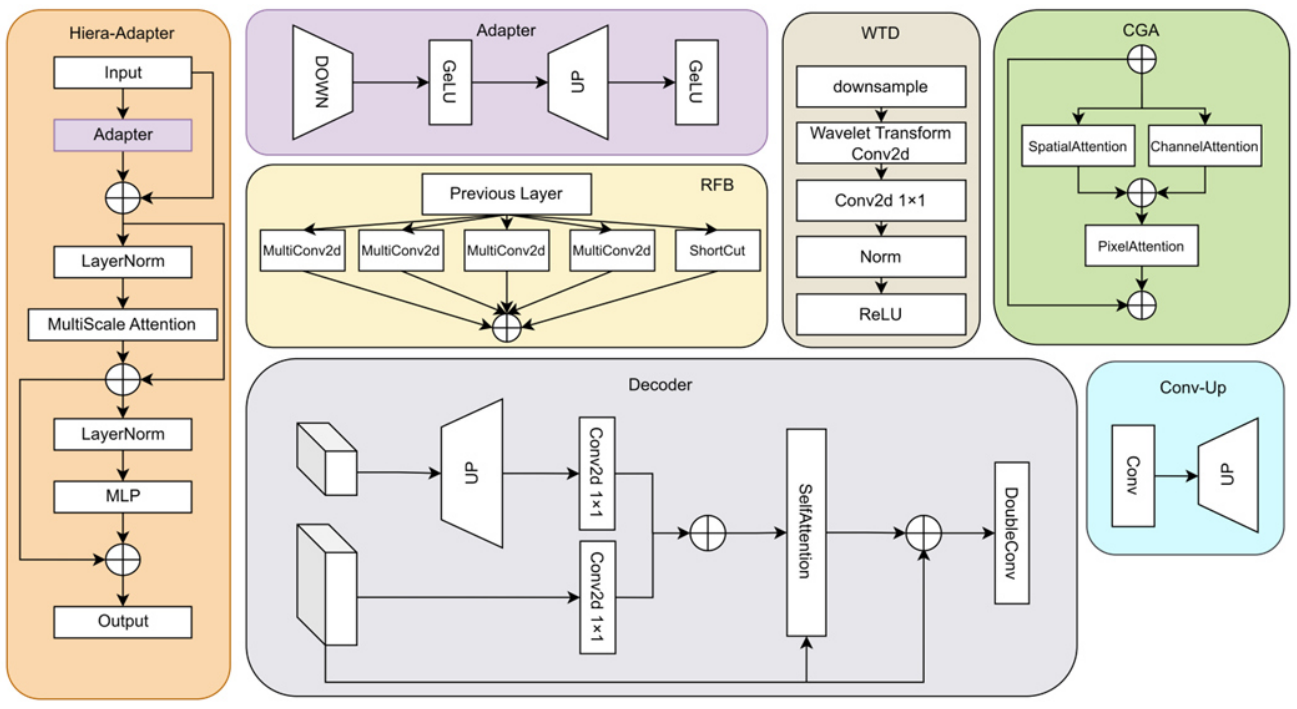

Fig. 4. DSU-Net Internal Module Detailed Diagrams

## IV. EXPERIMENT

### A. Datasets

To evaluate the performance of our model in salient object detection and camouflaged object detection, we trained and tested it on five datasets from salient object detection benchmark and four datasets from camouflaged object detection benchmark.

**Salient Object Detection (SOD)**. Salient Object Detection enables computer vision to mimic human perception by directing attention to prominent objects. Since its intro-duction in 2007, it has been widely applied in object recognition, tracking, and image segmentation. Our experiments utilize mainstream datasets, including DUTS[18], DUT-OMRON[19], HKU-IS[20], PASCAL-S[21], and ECSSD[22], as shown in Table I.

**Camouflaged Object Detection (COD)**. Camouflaged Object Detection involves accurately identifying targets that blend into their surrounding environment. For our experiments, we used CAMO[23], COD10K[24], CHAMELEON, and NC4K[25], as shown in Table II.

TABLE I. SALIENT OBJECT DETECTION DATASETS

| Dataset Name | Year | Training Set | Test Set |
|---|---|---|---|
| DUTS | 2017 | 10553 | 5019 |
| DUT-OMRON | 2013 | - | 5168 |
| HKU-IS | 2015 | - | 4447 |
| PASCAL-S | 2014 | - | 850 |
| ECSSD | 2013 | - | 1000 |

TABLE II. CAMOUFLAGE OBJECT DETECTION DATASET

| Dataset Name | Year | Training Set | Test Set |
|---|---|---|---|
| CAMO | 2019 | 1000 | 250 |
| COD10K | 2020 | 3040 | 2026 |
| CHAMELEON | 2018 | - | 76 |
| NC4K | 2021 | - | 4121 |

### B. Evaluation Metrics

To evaluate the model's prediction results, we follow the standards of salient object detection and camouflaged object detection benchmarks. We use several metrics, including the S-measure, F-measure, E-measure, and Mean Absolute Error (MAE), to assess the accuracy of the generated prediction maps. Below, we provide a detailed explanation of these common evaluation metrics.

**S-measure ($S_\alpha$)[26]**. To measure the structural similarity difference between the predicted map and the ground truth map, the formula is given by:

$$S_\alpha = \alpha S_o + (1 - \alpha) S_r \tag{3}$$

where α is the weight coefficient, ranging from [0,1], and is set to 0.5 according to benchmarking requirements, $S_o$ represents the object-oriented structural similarity metric, and $S_r$ represents the region-oriented structural similarity metric.

**F-measure ($F_\beta$)[27, 28]**. Combining precision and recall, the F-measure is used to comprehensively evaluate model performance. The formula is expressed as:

$$F = \frac{(1+\beta^2)\, P \cdot R}{\beta^2 \cdot P + R} \tag{4}$$

Here, $\beta^2$ serves as the weight to balance precision and recall. When $\beta^2 < 1$, it indicates that the evaluation places more emphasis on precision; when $\beta^2 > 1$, it signifies a greater emphasis on recall. P represents precision, which is the ratio of the number of correctly predicted positive samples to the total number of samples predicted as positive, while R denotes recall, which is the ratio of the number of correctly predicted positive samples to the total number of actual positive samples.

**E- measure ($E_\varphi$)[29]**. This metric is used to simulate human visual perception, evaluating both local and global similarity between the predicted image and the ground truth image at the pixel and image levels. The formula is defined as:

$$E_\varphi = \frac{1}{WH} \sum_{x=1}^{W} \sum_{y=1}^{H} \varphi\left( C(x,y) - G(x,y) \right) \tag{5}$$

Here, φ is an enhanced alignment matrix, W and H are the width and height of the input image, respectively, and C(x,y) and G(x,y) represent the pixel values at the (x,y) position in the predicted and ground truth images, respectively.

**Mean Absolute Error** (M)[30]. This metric measures the pixel-level accuracy difference between the predicted map and the ground truth annotation map, formulated as:

$$M = \frac{1}{WH} \sum_{x=1}^{W} \sum_{y=1}^{H} \left| C(x,y) - G(x,y) \right| \tag{6}$$

### C. Train and Test

During the training phase, the input images are resized to 352×352 and 518×518 respectively, for input into the Hiera and DINOv2 ViT encoders. The batch size is set to 8, and the number of training epochs is set to 50 following the benchmark testing conventions for salient object detection and camouflaged object detection. The learning rate is set to 0.001, with AdamW[31] as the chosen optimizer and a weight decay coefficient of 5e-4. Data augmentation is performed using random vertical and hori-zontal flips with a probability of 0.5. The code is developed using PyTorch, and the hardware resources utilized include an NVIDIA RTX 4090 with 24GB of video memory. For salient object detection, 10,553 images from the DUTS dataset are used as the training set, while the remaining 5,019 images and four other datasets (DUT-OMRON, HKU-IS, PASCAL-S, and ECSSD) serve as the test set. In camouflaged object detection, 1,000 images from CAMO and 3,040 images from COD10K are used as the training set, with the remaining 250 images from CAMO, 2,026 images from COD10K, and two additional datasets (CHAMELEON and NC4K) forming the test set. In summary, both types of object detection follow the standard training and testing partitioning schemes of their respective benchmark tests.

During the testing phase, we normalize the output of the last decoder to the range [0,1], multiply it by 255, and convert it to an 8-bit unsigned integer type to generate the output mask as a grayscale image.

### *D. Result*

The DSU-Net model achieves state-of-the-art performance in benchmark tests for salient object detection and camouflaged object detection, surpassing or matching the current SOTA methods.

The benchmark test results for salient object detection are shown in Table III and Table IV, where DSU-Net slightly outperforms SAM2-Unet across five datasets. In the benchmark test results for camouflaged object detection, as shown in Table V and Table VI, DSU-Net generally outperforms SAM2-Unet in four datasets, with performance improvements surpassing those observed in the salient object detection.

We compare our method with the latest SOTA model, SAM2-UNet, in both salient object detection and camouflaged object detection, with visual results presented in Fig. 4 and Fig. 5, respectively. Our method clearly excels in salient object detection, effectively suppressing background distractions and accurately segmenting the main object. In camouflaged object detection, it significantly outperforms SAM2-Unet, particularly in scenarios with subtle background differences and small object segmentation.

TABLE III. PERFORMANCE RESULTS OF SALIENT OBJECT DETECTION BENCHMARK TESTS ON DATASETS DUTS-TE, DUT-OMRON AND HKU-IS

| Methods | DUTS-TE | | | DUT-OMRON | | | HKU-IS | | |
|---|---|---|---|---|---|---|---|---|---|
| | $S_\alpha$ | $E_\varphi$ | $M$ | $S_\alpha$ | $E_\varphi$ | $M$ | $S_\alpha$ | $E_\varphi$ | $M$ |
| U2Net [32] | 0.874 | 0.884 | 0.044 | 0.847 | 0.872 | 0.054 | 0.916 | 0.948 | 0.031 |
| ICON[33] | 0.889 | 0.914 | 0.037 | 0.845 | 0.879 | 0.057 | 0.920 | 0.959 | 0.029 |
| EDN[34] | 0.892 | 0.925 | 0.035 | 0.850 | 0.877 | 0.049 | 0.924 | 0.955 | 0.026 |
| MENet [35] | 0.905 | 0.937 | 0.028 | 0.850 | 0.891 | 0.045 | 0.927 | 0.966 | 0.023 |
| SAM2-Unet[10] | 0.934 | 0.959 | 0.020 | 0.884 | 0.912 | 0.039 | 0.941 | **0.971** | 0.019 |
| DSU-Net(Ours) | **0.934** | **0.959** | **0.020** | **0.893** | **0.919** | **0.034** | **0.945** | 0.969 | **0.019** |

### *E. Ablation Study*

To assess the effectiveness of the features extracted by the DINOv2 ViT feature extractor when incorporated into the final-layer feature map of the SAM2 Hiera backbone network, and to demonstrate their role in facilitating the multi-scale feature collaboration and fusion in SAM2, we conducts ablation experiments, with the results presented in Table VII and Table VIII.

In this experiment, A represents DSU-Net without the DINOv2 ViT feature extraction module; B indicates a configuration where, instead of using the final-layer output feature map, DINOv2 ViT in DSU-Net extracts feature maps from four hierarchical Block blocks, which are separately injected into the four feature maps of SAM2 Hiera; C indicates that the output feature map of the final-layer of DINOv2 ViT in DSU-Net is injected into the all four feature maps of SAM2 Hiera.

TABLE IV. PERFORMANCE RESULTS OF SALIENT OBJECT DETECTION BENCHMARK TESTS ON DATASETS PASCAL-S AND ECSSD

| Methods | PASCAL-S | | | ECSSD | | |
|---|---|---|---|---|---|---|
| | $S_\alpha$ | $E_\varphi$ | $M$ | $S_\alpha$ | $E_\varphi$ | $M$ |
| U2Net[32] | 0.844 | 0.850 | 0.074 | 0.928 | 0.925 | 0.033 |
| ICON[33] | 0.861 | 0.893 | 0.064 | 0.929 | 0.954 | 0.032 |
| EDN[34] | 0.865 | 0.902 | 0.062 | 0.927 | 0.951 | 0.032 |
| MENet[35] | 0.872 | 0.913 | 0.054 | 0.928 | 0.954 | 0.031 |
| SAM2-Unet[10] | 0.894 | 0.931 | 0.043 | 0.950 | 0.970 | 0.020 |
| DSU-Net(Ours) | **0.898** | **0.935** | **0.042** | **0.953** | **0.972** | **0.018** |

TABLE V. PERFORMANCE RESULTS OF CAMOUFLAGED OBJECT DETECTION BENCHMARK TESTS ON DATASETS CHAMELEON AND CAMO

| Methods | CHAMELEON | | | | CAMO | | | |
|---|---|---|---|---|---|---|---|---|
| | $S_\alpha$ | $F_\beta$ | $E_\varphi$ | $M$ | $S_\alpha$ | $F_\beta$ | $E_\varphi$ | $M$ |
| SINet[24] | 0.872 | 0.832 | 0.936 | 0.034 | 0.745 | 0.712 | 0.804 | 0.092 |
| PFNet[36] | 0.882 | 0.820 | 0.931 | 0.033 | 0.782 | 0.751 | 0.841 | 0.085 |
| ZoomNet[37] | 0.902 | 0.858 | 0.943 | 0.024 | 0.820 | 0.792 | 0.877 | 0.066 |
| FEDER[38] | 0.903 | 0.856 | 0.947 | 0.026 | 0.836 | 0.807 | 0.897 | 0.066 |
| SAM2-Unet[10] | 0.914 | 0.863 | 0.961 | 0.022 | 0.884 | **0.861** | **0.932** | 0.042 |
| DSU-Net(Ours) | **0.923** | **0.869** | **0.963** | **0.020** | **0.889** | 0.859 | 0.931 | **0.041** |

TABLE VI. PERFORMANCE RESULTS OF CAMOUFLAGED OBJECT DETECTION BENCHMARK TESTS ON DATASETS COD10K AND NC4K

| Methods | COD10K | | | | NC4K | | | |
|---|---|---|---|---|---|---|---|---|
| | $S_\alpha$ | $F_\beta$ | $E_\varphi$ | $M$ | $S_\alpha$ | $F_\beta$ | $E_\varphi$ | $M$ |
| SINet[24] | 0.776 | 0.667 | 0.864 | 0.043 | 0.808 | 0.768 | 0.871 | 0.058 |
| PFNet[36] | 0.800 | 0.676 | 0.877 | 0.040 | 0.829 | 0.779 | 0.887 | 0.053 |
| ZoomNet[37] | 0.838 | 0.740 | 0.888 | 0.029 | 0.853 | 0.814 | 0.896 | 0.043 |
| FEDER[38] | 0.844 | 0.784 | 0.911 | 0.029 | 0.862 | 0.824 | 0.913 | 0.042 |
| SAM2-Unet[10] | 0.880 | 0.789 | 0.936 | 0.021 | 0.901 | 0.863 | 0.941 | 0.029 |
| DSU-Net(Ours) | **0.891** | **0.819** | **0.947** | **0.019** | **0.909** | **0.868** | **0.948** | **0.026** |

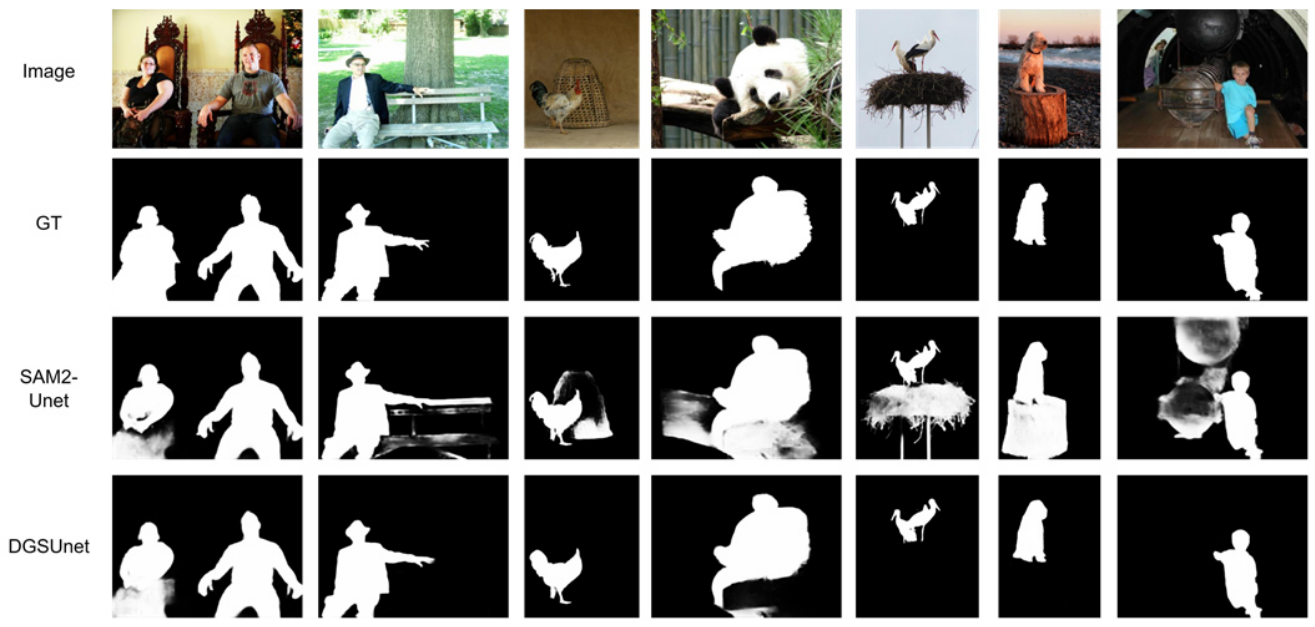

Fig. 5. Camouflaged Object Detection Visualization Result

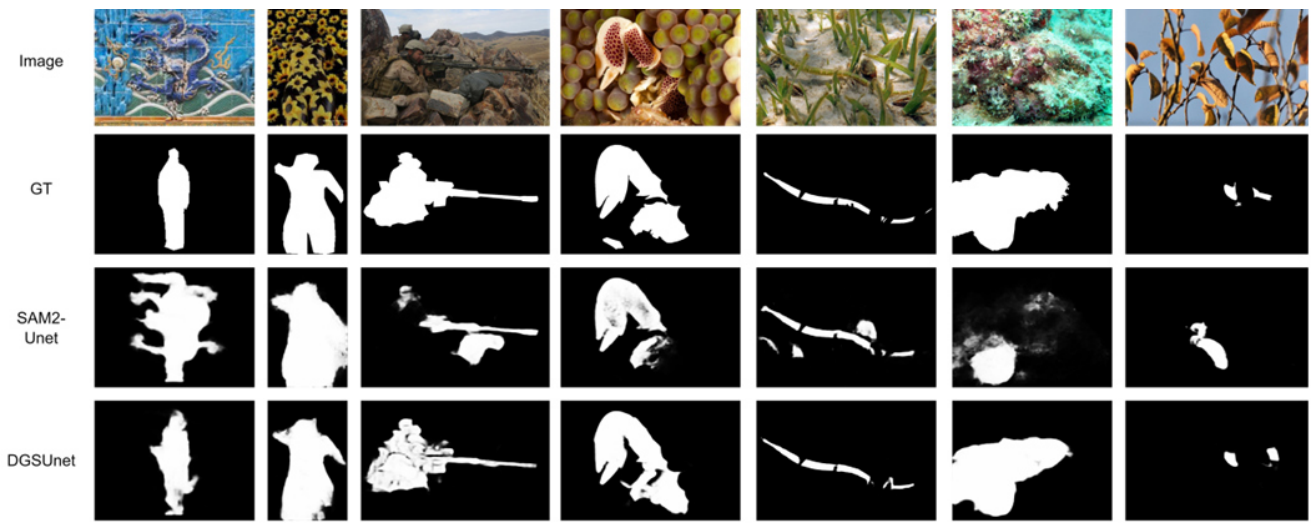

Fig. 6. Salient Object Detection Visualization Results.

TABLE VII. ABLATION STUDY RESULT OF SALIENT OBJECT DETECTION BENCHMARK TESTS ON DATASETS DUTS-TE, DUT-OMRON

| Methods | DUTS-TE | | | DUT-OMRON | | |
|---|---|---|---|---|---|---|
| | $S_\alpha$ | $E_\varphi$ | $M$ | $S_\alpha$ | $E_\varphi$ | $M$ |
| A | 0.931 | 0.955 | 0.023 | 0.880 | 0.909 | 0.041 |
| B | 0.926 | 0.950 | 0.025 | 0.879 | 0.907 | 0.044 |
| C | 0.928 | 0.952 | 0.024 | 0.880 | 0.907 | 0.042 |
| DSU-Net(Ours) | **0.934** | **0.959** | **0.020** | **0.893** | **0.919** | **0.034** |

TABLE VIII. ABLATION STUDY RESULTS OF CAMOUFLAGED OBJECT DETECTION BENCHMARK TESTS ON DATASETS CAMO AND COD10K

| Methods | CAMO | | | | COD10K | | | |
|---|---|---|---|---|---|---|---|---|
| | $S\alpha$ | $F\beta$ | $E\varphi$ | $M$ | $S\alpha$ | $F\beta$ | $E\varphi$ | $M$ |
| A | 0.881 | 0.852 | 0.924 | 0.045 | 0.879 | 0.800 | 0.938 | 0.020 |
| B | 0.875 | 0.848 | 0.919 | 0.048 | 0.868 | 0.792 | 0.933 | 0.022 |
| C | 0.880 | 0.851 | 0.922 | 0.043 | 0.881 | 0.801 | 0.939 | 0.020 |
| DSU-Net(Ours) | **0.889** | **0.859** | **0.931** | **0.041** | **0.891** | **0.819** | **0.947** | **0.019** |

Clearly, the fusion of features extracted exclusively from the final-layer of DINOv2 ViT with the feature map from the last layer of SAM2 Hiera proves to be the most effective.

## V. CONCLUSION

We propose a novel architecture called DSU-Net, which features a dual-encoder collaborative U-Net designed to improve segmentation accuracy in specialized fields. The performance enhancement of DSU-Net is attributed to its self-supervised semantic guidance mechanism, an improved feature fusion module, and the incorporation of feature adapters in Hiera. This architecture aims to address two key challenges faced by U-Net: limited domain feature representation capability, and high training costs. We evaluated DSU-Net on five datasets from significant object detection benchmarks and four datasets from camouflaged object detection benchmarks, demonstrating its clear advantages in image segmentation compared to various advanced architectures and frameworks. It achieved exceptional performance surpassing single-encoder models, and is applicable to a wide range of image segmentation tasks.

Although our method is effective, it does come with a considerable number of model parameters, which poses challenges for lightweight deployment and practical applications in production environments. Moving forward, we will explore solutions such as model pruning and distillation learning to develop models with fewer parameters while maintaining superior performance.

## Authors' Background

| Name | Email | Position (Prof , Assoc. Prof. etc.) | Research Field | Homepage URL |
|---|---|---|---|---|
| Yimin Xu | xuyimin23@mails.ucas.ac.cn | Graduate Student | Computer Vision | https://github.com/CheneyXuYiMin |
| Fan Yang | yangfan@casit.com.cn | Dr. | Artificial Intelligence | https://people.ucas.ac.cn/~0069345 |
| Bin Xu | xubin23@mails.ucas.ac.cn | Graduate Student | Natural Language Processing | https://github.com/zhihh |
| | | | | |
| | | | | |
| | | | | |
| | | | | |